\documentclass{article} 
\usepackage{iclr2026_conference,times}

\usepackage{graphicx}
\usepackage{booktabs}
\usepackage{titletoc}
\usepackage[utf8]{inputenc} 
\usepackage[T1]{fontenc}    
\usepackage{url}            
\usepackage{booktabs}       
\usepackage{amsfonts}       
\usepackage{nicefrac}       
\usepackage{microtype}      
\usepackage{xcolor}         
\usepackage{authblk}        
\usepackage{graphicx}
\definecolor{linkcolor}{HTML}{c0392b}
\iclrfinalcopy
\usepackage{amsmath} 
\usepackage{titletoc}

\usepackage{amsmath,amsfonts,bm}

\def\eqref#1{equation~\ref{#1}}

\def\1{\bm{1}}

\DeclareMathAlphabet{\mathsfit}{\encodingdefault}{\sfdefault}{m}{sl}
\SetMathAlphabet{\mathsfit}{bold}{\encodingdefault}{\sfdefault}{bx}{n}

\usepackage{multirow}
\usepackage{makecell}
\usepackage{enumitem}
\usepackage[table]{xcolor} 
\definecolor{rowgray}{gray}{0.9} 
\definecolor{lightblue}{HTML}{E6F2FF}
\definecolor{MWOrange}{HTML}{FF7F0E}
\definecolor{MWGreen}{HTML}{32CD32}
\colorlet{lightorange}{MWOrange!12}
\colorlet{lightgreen}{MWGreen!10}
\usepackage{caption}
\usepackage{hyperref}
\usepackage{url}
\usepackage{amssymb}

\title{Reinforcement Learning with Inverse Rewards for World Model Post-training}

\author{%
  \textbf{Yang Ye}\textsuperscript{1} \quad
  \textbf{Tianyu He}\textsuperscript{1}\quad
  \textbf{Shuo Yang}\quad
  \textbf{Jiang Bian}\textsuperscript{1}
}

\affil{
  \textsuperscript{1} Microsoft Research
}

\begin{document}

\maketitle
\begin{abstract}
World models simulate dynamic environments, enabling agents to interact with diverse input modalities. Although recent advances have improved the visual quality and temporal consistency of video world models, their ability of accurately modeling human-specified actions remains underexplored. 
Reinforcement learning presents a promising approach for directly improving the suboptimal action-following capability of pre-trained models, assuming that an appropriate reward function can be defined.
However, transferring reinforcement learning post-training methods to world model is impractical due to the prohibitive cost of large-scale preference annotations and the infeasibility of constructing rule-based video verifiers. 
To address this gap, we propose Reinforcement Learning with Inverse Rewards (RLIR), a post-training framework that derives verifiable reward signals by recovering input actions from generated videos using an Inverse Dynamics Model. By mapping high-dimensional video modality to a low-dimensional action space, RLIR provides an objective and verifiable reward for optimization via Group Relative Policy Optimization. Experiments across autoregressive and diffusion paradigms demonstrate 5–10\% gains in action-following, up to 10\% improvements in visual quality, and higher human preference scores, establishing RLIR as the first post-training method specifically designed to enhance action-following in video world models.
\end{abstract}

\section{Introduction}
\label{sec:introduction}

World models aim to simulate dynamic environments, enabling intelligent agents to effectively interact with various input modalities such as robot actions, camera poses, or keyboard commands~\citep{worldmodel}. Building on recent advances in generative modeling~\citep{ho2020denoising,rombach2022high} and large-scale video datasets, contemporary video world models achieve substantial improvements in both fidelity and diversity of synthesized visual environments through training action-conditioned video generation models~\citep{hu2023gaia,mineworld,bar2025navigation}.

To function as reliable simulators, world models must satisfy three key requirements: producing high-fidelity visual content, maintaining long-horizon temporal consistency, and accurately following human-specified actions. While extensive research has addressed the first two challenges~\citep{genie3}, with approaches such as extending context windows~\citep{liu2024world,zhang2025packing,gu2025long} and incorporating 3D priors~\citep{wu2025video,wu2025geometry,xiao2025worldmem}, the problem of accurate action-following remains comparatively underexplored~\citep{trajfollow}, despite its central role in controllable and interactive world modeling.

In natural language processing, reinforcement learning-based post-training has proven highly effective for aligning large language models with human preferences~\citep{rlhf} and for enhancing reasoning capabilities~\citep{rlvr}. These successes suggest reinforcement post-training as a promising direction for video world models. However, direct transfer of existing techniques faces critical obstacles: (1) collecting human preference annotations at scale for video data is prohibitively expensive and prone to bias, and (2) while approaches such as RLVR~\citep{rlvr} mitigate this issue by leveraging faithful, rule-based rewards and often achieve strong performance, their applicability remains limited to narrow domains (e.g., coding and mathematics). In particular, designing rule-based verifiers to reliably assess the quality of generated video is generally infeasible.

To overcome these challenges, we introduce Reinforcement Learning with Inverse Rewards (RLIR), a post-training framework for world models. The core idea is that, \textit{rather than evaluating model output directly in the high-dimensional video space, RLIR derives reward signals in the low-dimensional input space (e.g., actions) by employing an inverse model that predicts conditioning actions from the generated videos}. 
Within our post training framework, we begin by employing either autoregressive or diffusion models to generate video sequences conditioned on input actions. An Inverse Dynamics Model (IDM) is then utilized to translate actions back from the generated videos. Given access to the ground-truth input actions, we can compare the inferred actions with the original input actions on a per-frame basis, thereby obtaining a verifiable reward signal. The reward increases with the degree of alignment between the inferred and ground-truth actions. Finally, we adopt the Group Relative Policy Optimization (GRPO) algorithm~\citep{deepseekmath} to optimize the world model according to the relative advantages of the generated sequences.

Our approach is grounded in the key insight that, although multiple valid videos may correspond to the same action sequence, all high-quality generations must faithfully encode the input actions. Deviations such as temporal inconsistencies or visual artifacts reduce IDM accuracy, thereby naturally penalizing inferior outputs. Compared with human preference-based rewards, this action-consistency signal offers a more objective, scalable, and low-bias criterion for post-training world models.

We evaluate RLIR on interactive game generation domain across both autoregressive (next-token prediction) and diffusion world models. The results corroborate our key insight, demonstrating consistent improvements in action-following accuracy and visual quality across different generative paradigms. 
In summary, our contributions are threefold:

\textbf{i)} We introduce Reinforcement Learning with Inverse Rewards, a post-training paradigm that employs an inverse model to map inherently unverifiable video outputs into a verifiable, low-dimensional action sequence, thereby enabling reinforcement post-training to video world models.

\textbf{ii)} We leverage RLIR to improve action-following ability in world models, demonstrating its remarkable effectiveness across both autoregressive and diffusion paradigms. To the best of our knowledge, this is the first post-training method specifically designed to improve action-following ability.

\textbf{iii)} Experiments on both generative paradigms show consistent 5\%-10\% gains on action-following metrics and up to a 10\% improvement in visual quality, with superior human preference scores.

\section{Related Work}\label{sec: related_work}
\subsection{World Model}
World models~\citep{hunyuanworld,dinowm,cosmos} are generative systems that enable agents or humans to effectively interact with dynamic environments. 
Leveraging recent progress in generative modeling~\citep{opensoraplan,consisid} and the availability of large-scale datasets~\citep{panda70m,imgedit,opens2v}, modern video world models have significantly enhanced the fidelity and diversity of synthesized visual environments by training action-conditioned video generation models.
In the gaming domain, numerous studies~\citep{mineworld,diagd,gamefactory} simulate interactive video games as well as real-world exploration, further extending the controllability and scalability of world models. 
Prior work has emphasized visual quality~\citep{genie3} and long-horizon physical consistency~\citep{xiao2025worldmem,wu2025video} in world models, yet the issue of inaccurate action-following remains unaddressed. Our paper focuses on improving the action-following capability of world models.

\subsection{Reinforcement Learning for Generative Models}
Reinforcement learning~\citep{openaio1,lookback} has emerged as a critical paradigm for post-training to better align with human preferences or task-specific objectives. DeepSeek-R1~\citep{deepseekr1} introduces verifiable rewards and uses group relative policy optimization (GRPO) as its training method, which is more memory efficient by removing the need for a value network. Recently, GRPO-style methods~\citep{flowgrpo,dancegrpo} have progressed rapidly in generative models. However, their reward functions primarily rely on metrics such as aesthetics~\citep{aespredictor}, text-image alignment~\citep{clip}, or use Multimodal-Large-Language Model as a judger~\citep{videollava,moellava}, which are constrained by the accuracy and biases of the reward models and often result in suboptimal performance. In our work, we use action accuracy as a reward, an objective criterion that can be precisely measured via an Inverse Dynamics Model, which is the first post-training method designed for improving action-following in video world models.

\section{Preliminaries}\label{sec:preliminaries}

\subsection{Inverse Dynamics Model}
Given a trajectory of $T$ observations $o_t: t \in [1...T]$, an Inverse Dynamics Model (IDM) estimates the action that transitions $o_t$ to $o_{t+1}$; formally, it models $p_{\textrm{\textsubscript{IDM}}}(a_t | o_t, o_{t+1})$. The IDM is trained on a contractor-labeled dataset by minimizing the negative log-likelihood of the ground-truth action at time $t$ given $(o_t, o_{t+1})$. Since the model leverages information from all video frames (including both past and future observations) to infer the current action, and given that the action space is substantially lower-dimensional than the raw video space, the IDM can achieve accurate prediction even with a limited amount of labeled data. The effectiveness of IDMs has been extensively validated in various domains, including robotic manipulation~\citep{unipi,susie,anypos}, game environments~\citep{vpt}, and 3D geometric perception~\citep{vipe}.

\begin{figure*}[!thbp]
    \centering
    \includegraphics[width=1\linewidth]{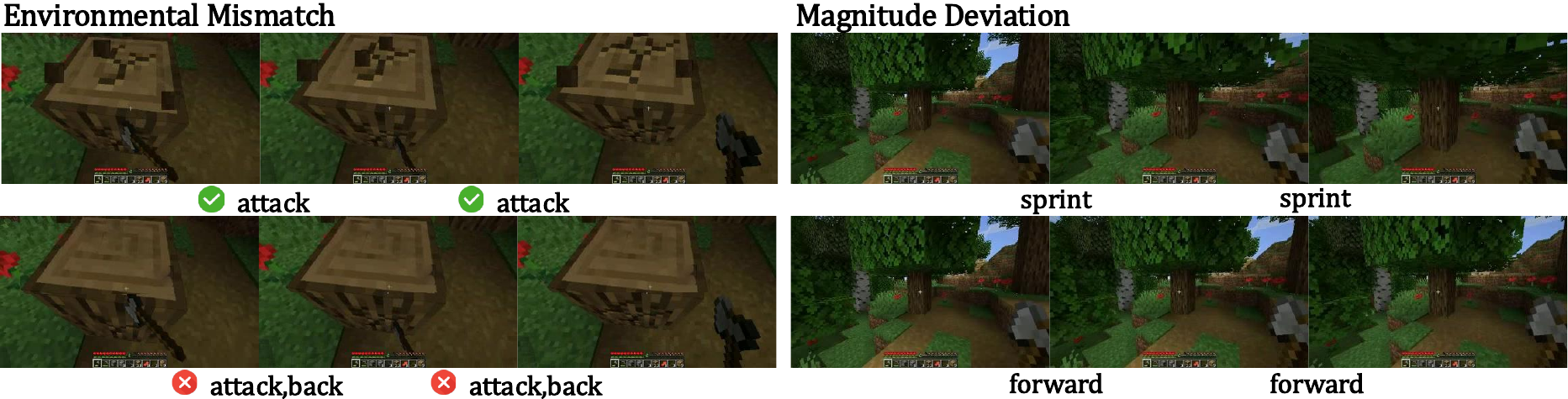}
    \caption{Inverse Dynamics Model (IDM) is highly sensitive to subtle environmental changes and action magnitudes. \textbf{(left)} The IDM flags the failure to produce cracks on the trunk as the action `attack,back' rather than the ground-truth `attack' and therefore labels it as a negative sample. \textbf{(right)} The IDM detects even subtle differences in action magnitudes (e.g., `forward' and `sprint').}
    \label{fig: idmcase}
\end{figure*}

A well-trained Inverse Dynamics Model exhibits high sensitivity to minor visual artifacts and subtle variations of actions. In Figure~\ref{fig: idmcase} (left), we manually retouch only the cracks of the trunk to emulate a localized failure in world model generation. The IDM detects the inconsistency and consequently outputs an incorrect action prediction. As shown on the right, the IDM can also reliably discriminate between actions such as `forward' and `sprint', even when the visual differences are minimal. Prior work~\citep{vpt} further demonstrates the effectiveness of IDM in the Minecraft environment, reporting 90.6\% accuracy on keypress prediction and an $R^2$ of 0.97 for mouse movement regression.

\subsection{Group Relative Policy Optimization}

Group Relative Policy Optimization (GRPO)~\citep{deepseekmath,deepseekr1} is originally developed for post-training LLMs with reinforcement learning. Compared to Proximal Policy
Optimization~\citep{ppo}, GRPO dispenses with a value function and estimates advantages in a group-relative manner. Specifically, given a question $q$, it samples a set of $G$ responses $\{o_i\}_{i=1}^G$ from the behavior policy $p_{\theta_\text{old}}$, and computes the advantage of each response by normalizing its reward $R_i$ within the group:
\begin{equation}
\hat{A}_{i,t} = \frac{R_i-\operatorname{mean}(\{R_i\}_{i=1}^G)}{\operatorname{std}(\{R_i\}_{i=1}^G)}
\end{equation}
Similar to PPO, GRPO uses a clipped objective with a KL divergence~\citep{kl_div} penalty:
\begin{equation}
\label{eq:grpoloss}
\begin{aligned}
&\mathcal{J}_\text{GRPO}(\theta) = \mathbb{E}_{q\sim \mathcal{D}, \{o_i\}_{i=1}^G\sim p_{\theta_\text{old}}(\cdot\mid q)} \\&
=\Bigg[ \frac{1}{G}\sum_{i=1}^{G} \frac{1}{|o_i|}\sum_{t=1}^{|o_i|} \Bigg(
\min \Big( \frac{p_{\theta}^{i,t}}{p_{\theta_{\text{old}}}^{i,t}} \hat{A}_{i,t},  
\ \text{clip} \Big( \frac{p_{\theta}^{i,t}}{p_{\theta_{\text{old}}}^{i,t}}, 1 - \varepsilon, 1 + \varepsilon \Big) \hat{A}_{i,t} \Big)
- \beta D_{\text{KL}}\left[p_{\theta} || p_{\text{ref}}\right] 
\Bigg) \Bigg],
\end{aligned}
\end{equation}
where $p_{\theta}^{i,t}$ denotes $p_{\theta}(o_{i,t} \mid q, o_{i,<t})$ for simplicity. Numerous recent works~\citep{dapo,gspo,gfpo} optimize GRPO for algorithmic efficiency or performance. For simplicity, we adopt the vanilla GRPO algorithm in this paper.

\section{Method}\label{sec: method}

In this section, we describe our method in detail. We first briefly introduce the problem and motivations. Then in Section~\ref{subsec:idm_as_reward}, we describe how an Inverse Dynamics Model (IDM) is used as the reward model in Reinforcement Learning with Inverse Reward (RLIR). Sections~\ref{subsec:mineworld} and~\ref{subsec:nfd} demonstrate the application of RLIR to two representative classes of world models, namely autoregressive world model and diffusion world model, respectively.

\paragraph{Problem Definition}

To provide a simplified description of world models, we denote the model-generated frames as $\hat{x}$. Given an initial state $x_0$ and an action sequence $a_1, \ldots, a_n$, the world model generates each frame $\hat{x}_i$ conditioned on the initial state $x_0$, the previously generated frames $\hat{x}_1, \ldots, \hat{x}_{i-1}$ and the corresponding actions $a_1, \ldots, a_i$. 

\paragraph{Motivation}

To ensure that world models accurately follow human-specified actions, we focus on enhancing their action-following capability. Our approach is based on the insight that if the generated video frames faithfully reflect the input actions, then these actions can be reliably recovered from the generated frames. Guided by this insight, we initialize with a pretrained video world model and incorporate an IDM as a reward function to enhance action alignment through reinforcement learning.

\subsection{IDM as Reward Model}\label{subsec:idm_as_reward}

\begin{figure*}[t]
    \centering
    \includegraphics[width=0.98\linewidth]{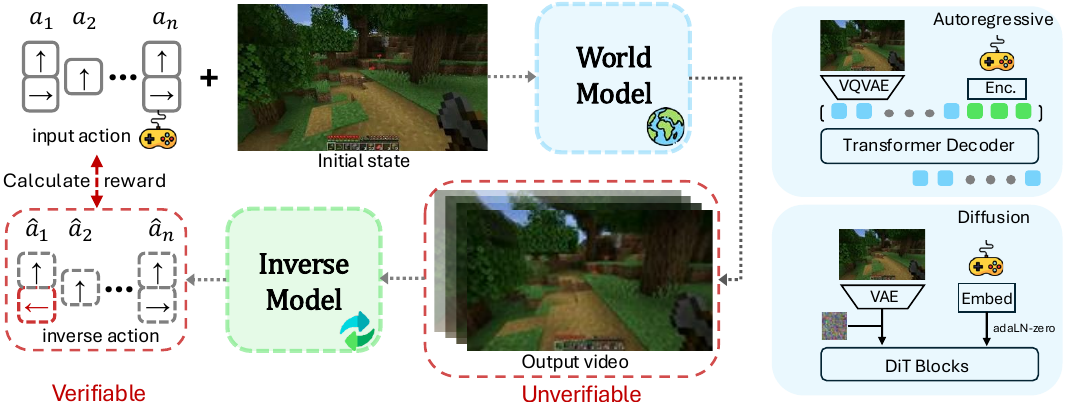}
    \caption{\textbf{Overview of RLIR.} Given the input actions, the world model generates video sequences conditioned on the input actions. An Inverse Dynamics Model (IDM) is then utilized to derive verifiable reward signals by recovering input actions from generated videos. We adopt Group Relative Policy Optimization (GRPO) to optimize the world model according to the alignment between the inferred and ground-truth input actions. We validate RLIR on both autoregressive and diffusion world models; architectures are shown on the right.}
    \label{fig: algorithm}
\end{figure*}

Research on applying reinforcement learning to world models remains relatively limited. Existing approaches~\citep{videoalign,hps} primarily assess visual quality, but they cannot reliably determine whether the intended actions have been correctly executed at the frame level, and they are susceptible to bias. Assigning rewards becomes substantially easier if we map video back to the action space and evaluate rewards in the action space. Specifically, while a world model maps an action sequence to video, by projecting the generated video back into the action space and comparing it to the original actions, we obtain a direct measure of the model’s action-following fidelity. This video-to-action mapping can be implemented simply and accurately using an IDM. Formally, for each generated trajectory $T_j = [x_0, \hat{x}_1, \ldots, \hat{x}_n]$, there exists a corresponding ground truth action sequence $a_1, \ldots, a_n$. We employ a well-trained IDM that takes the generated trajectory $T_j$ as input and predicts the actions $\hat{a}_1, \ldots, \hat{a}_n$. Given that the IDM has been thoroughly trained to predict actions with high precision, any discrepancy between the predicted actions $\hat{a}_i$ and the ground truth actions $a_i$ can be attributed to errors in the world model. Our reward function can thus be formalized as follows:
\begin{equation}
    R_{T_j} = \frac{1}{n} \sum_{i=1}^{n} r(\hat{a}_{i}, {a}_{i}), \quad \quad r(\hat{a}_{i}, {a}_{i}) \;\triangleq
      \begin{cases}
        1, & \text{if } \hat{a}_{i} = {a}_{i}, \\
        0, & \text{otherwise}.
      \end{cases}
\label{eq: reward}
\end{equation}
Unlike previous reward models for video generation, IDM is trained exclusively on videos with ground-truth action annotations, introducing no additional bias and yielding precise action estimates that translate directly into a verifiable reward signal.

\subsection{Autoregressive World Model}\label{subsec:mineworld}

An autoregressive world model generates a video sequence by iteratively predicting the next visual token in the sequence. We utilize the pretrained MineWorld~\citep{mineworld} as our baseline model. MineWorld employs a visual-action autoregressive Transformer that takes pairs of game scenes and corresponding actions as input and generates subsequent scenes conditioned on the actions. The inputs comprise two modalities: gameplay videos represented as a sequence of states $x_i$ and actions $a_i$ captured from mouse and keyboard events. 
For each state-action pair $(x_i, a_i)$, a VQ-VAE~\citep{vqvae} tokenizer encodes $x_i$ into a sequence of quantized codes $t$, and an action tokenizer encodes $a_i$ into a flat sequence of discrete tokens separately. The tokenized data are structured as follows:
\begin{equation}
    (t^{x_i}_0, \cdots, t^{x_i}_n, [\texttt{aBOS}], t^{a_i}_0, \cdots, t^{a_i}_8, \texttt{[aEOS]}).
\end{equation}
The Transformer architecture follows LLaMA~\citep{llama3} and treats tokens that represent game states and actions equally. The model is trained with next-token prediction to learn rich representations of game states and to model dependencies between states and actions.

Our post-training method is based on Group Relative Policy Optimization. In MineWorld, the rollout sequence comprises visual tokens and action tokens, and the latter are derived from external inputs. Optimizing visual tokens improves action-following ability and generative performance, but applying the same optimization to action tokens can induce undesirable training dynamics. During training, we address this challenge by implementing loss masking for action tokens, effectively disregarding the loss associated with these tokens. This ensures that the policy-gradient objective is computed solely on tokens generated by the world model, excluding action tokens from optimization.

\subsection{Diffusion World Model}\label{subsec:nfd}
In contrast to the autoregressive world model, the diffusion world model leverages Diffusion Forcing~\citep{diffusionforcing} to generate videos by autoregressively denoising future frames. We use the pretrained NFD~\citep{NFD} as our baseline model. NFD uses diffusion Transformer blocks with block-wise causal attention. During inference, it can perform causal sampling across frames while applying parallel diffusion denoising to all visual tokens within each frame. NFD uses an image-level tokenizer to convert each frame into a sequence of tokens in a continuous space. For action processing, it uses a linear layer to map actions to vector embeddings, and AdaLN-Zero~\citep{peebles2023scalable} conditioning injects action information into the model.

Inspired by prior work~\citep{dancegrpo,flowgrpo}, we convert the deterministic Flow-ODE used in NFD into an equivalent SDE whose marginal probability density matches that of the original model at all timesteps. Within the diffusion framework, the denoising dynamics of diffusion models can be cast as a Markov decision process.

\begin{equation}
  \begin{alignedat}{2}
    \mathbf{s}_t \;&\triangleq (\mathbf{c}, t, \mathbf{z}_t), \qquad
    & \pi(\mathbf{a}_t \mid \mathbf{s}_t) \;&\triangleq p(\mathbf{z}_{t-1} \mid \mathbf{z}_t, \mathbf{c}), \\
    P(\mathbf{s}_{t+1} \mid \mathbf{s}_t, \mathbf{a}_t) \;&\triangleq (\delta_{\mathbf{c}}, \delta_{t-1}, \delta_{\mathbf{z}_{t-1}}), \qquad
    & R(\mathbf{s}_t, \mathbf{a}_t) \;&\triangleq
      \begin{cases}
        r(\mathbf{z}_0, \mathbf{c}), & \text{if } t = 0, \\
        0, & \text{otherwise}.
      \end{cases}
  \end{alignedat}
\end{equation}

In the formulation, $\mathbf{c}$ denotes the action-conditioning input, and $\pi(\mathbf{a}_t\mid\mathbf{s}_t)$ represents the transition probability from the latent state $z_t$ to $z_{t-1}$. Each trajectory consists of $T$ timesteps, after which the transition function $P$ leads to a terminal state. The detailed reward $r(\mathbf{z}_0,\mathbf{c})$ is given in Equation~\ref{eq: reward}. We provide reward only at $t=0$ for the final output, with no reward at any other timestep.

\section{Experiments}\label{sec: experiments}
\subsection{Setups}
\paragraph{Implementation Detail}
We use the VPT dataset~\citep{vpt} for post-training. We apply a preprocessing pipeline that removes data that cannot be processed by the Inverse Dynamics Model (IDM), specifically frames recorded during GUI interactions or those in which the scene is static. All visual inputs are resized to $384 \times 224$ pixels. In practice, about 1,000 training samples are sufficient for the model to converge.

All training is conducted on AMD MI300X GPUs. In the post-training stage, we load the pretrained weights of MineWorld~\citep{mineworld} and NFD~\citep{NFD} for corresponding experiments. For the IDM, we use the VPT-pretrained model~\citep{vpt}, trained on 2,000 hours of carefully curated gameplay videos. We observe that the prediction accuracy of IDM increases with video length. Therefore, we set the inference length to 16 frames during training. Additional hyperparameter settings are provided in Appendix~\ref{appendix:impl_detail}.

In evaluation stage, all hyperparameters are kept identical to baseline settings. For MineWorld, we set Top-$p$ sampling to 0.8. For NFD, we use DPM-Solver++~\citep{dpm} with 18 sample steps.

\paragraph{Evaluation Protocol}
Evaluation proceeds as follows: given an initial frame and a sequence of 15 actions, the model predicts the next frame at each step conditioned on the action associated with the current frame, producing a 16-frame video that we assess for video quality and action following. For video quality, we report Fréchet Video Distance (FVD)~\citep{FVD}, PSNR~\citep{psnr} and VBench~\citep{vbench}, which measure dynamics and visual quality. For action following, we adopt the MineWorld evaluation protocol and use the IDM to infer actions from videos. We report F1, precision, and recall scores to evaluate the action classification accuracy. We use the official data split from MineWorld~\citep{mineworld}.
To ensure comparability with prior work, we use the results reported in MineWorld and NFD as baselines.
More details are listed in Appendix~\ref{appendix:impl_detail}.
We also conduct human evaluations to validate that these metrics align with human preferences.

\subsection{Main Results} 
As shown in Table~\ref{tab: main_result}, the post-trained models significantly outperform the baselines across different paradigms and parameter scales, yielding substantial improvements in action-following accuracy, as reflected by higher F1, recall, and precision scores for actions. Moreover, visual quality metrics such as FVD, PSNR and image quality from VBench also show improvements relative to the baselines. We also report the action-following metric of ground truth videos in the table, which represents the upper bound of the IDM’s accuracy and therefore serves as the theoretical upper limit for RLIR performance. After post-training, the model nearly attains this bound.

\begin{table}[thbp]
    \centering
    \caption{Comparison of results with and without RLIR post-training across two model architectures and diverse parameter scales. RLIR consistently improves action-following ability and visual quality. ``GT'' denotes ground truth videos. ``Img. Qual.'' is short for ``image quality''.}
    \vspace{1mm}
    \label{tab: main_result}
    \resizebox{\textwidth}{!}{
        \begin{tabular}{cl|ccccccc}
            \toprule
            Model & Param.  & F1$\uparrow$       & Recall$\uparrow$  & Precision$\uparrow$  & FVD$\downarrow$        & PSNR$\uparrow$      & Img. Qual.$\uparrow$  & Dynamic\\ 
            \midrule
            \multirow{6}{*}{\textbf{\makecell[c]{Mine-\\World}}} & 300M     & 0.70       & 0.71      & 0.72      & 246       & 15.13     & 0.675  & 0.97     \\
            
            & \cellcolor{lightblue} w/ RLIR   & \cellcolor{lightblue}0.77     & \cellcolor{lightblue}0.76      & \cellcolor{lightblue}0.79     & \cellcolor{lightblue}231    & \cellcolor{lightblue}15.58          & \cellcolor{lightblue}0.672 &   \cellcolor{lightblue}0.97    \\
            \cmidrule(l){2-9}
            & 700M     & 0.70       & 0.71      & 0.72      & 231       & 15.32  & 0.677     & 0.96              \\
            & \cellcolor{lightblue} w/ RLIR   & \cellcolor{lightblue}\textbf{0.81}     & \cellcolor{lightblue}0.80       & \cellcolor{lightblue}\textbf{0.84}     & \cellcolor{lightblue}207     & \cellcolor{lightblue}15.78       & \cellcolor{lightblue}0.678 & \cellcolor{lightblue}0.97        \\
            \cmidrule(l){2-9}
            & 1200M    & 0.76      & 0.73      & 0.73      & 227       & 15.69      &  0.682 &  0.97          \\
            & \cellcolor{lightblue} w/ RLIR  &\cellcolor{lightblue} \textbf{0.81}      & \cellcolor{lightblue}\textbf{0.81}     & \cellcolor{lightblue}0.83   & \cellcolor{lightblue}\textbf{205}     & \cellcolor{lightblue}\textbf{15.99}       &  \cellcolor{lightblue}\textbf{0.684}  & \cellcolor{lightblue}0.96          \\ 
            \midrule[0.65pt]
            \multirow{4}{*}{\textbf{NFD}} & 310M   & 0.69     & 0.69      & 0.71     & 212    & 16.46          &   0.678  &  1.00       \\
            & \cellcolor{lightblue} w/ RLIR   &  \cellcolor{lightblue} 0.76       &  \cellcolor{lightblue}0.76        &   \cellcolor{lightblue}0.77      &  \cellcolor{lightblue}195      &  \cellcolor{lightblue}17.38     &  \cellcolor{lightblue}0.687   &  \cellcolor{lightblue}0.99   \\
            \cmidrule(l){2-9}
            & 774M   & 0.77     & 0.78       & 0.78     & 184     & 16.95   &\textbf{0.692}  &  0.99 \\
            & \cellcolor{lightblue} w/ RLIR   & \cellcolor{lightblue}\textbf{0.83}      &  \cellcolor{lightblue}\textbf{0.83}     &    \cellcolor{lightblue}\textbf{0.85}     &  \cellcolor{lightblue}\textbf{180}   & \cellcolor{lightblue}\textbf{17.48}         & \cellcolor{lightblue}0.688 &  \cellcolor{lightblue}1.00      \\
            \midrule[0.65pt]
            \textbf{GT} & ----------- & 0.87 & 0.86 & 0.88 & --- & ------ & 0.704 & 1.00 \\
            \bottomrule
        \end{tabular}
    }
\end{table}

\begin{figure*}[!htbp]
    \centering
    \includegraphics[width=1\linewidth]{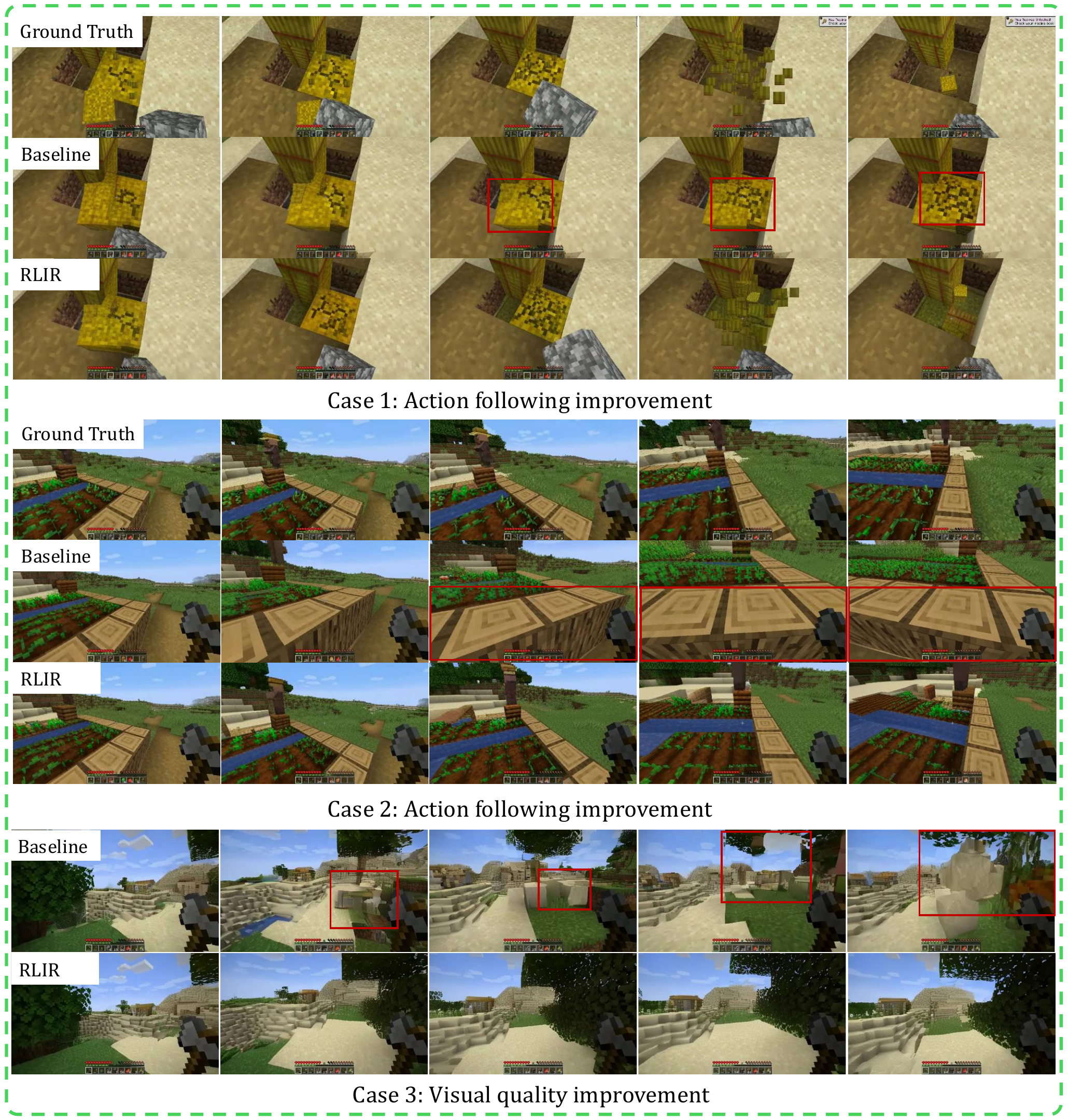}
    \caption{Qualitative comparison between RLIR and baseline. The figure shows the ground truth, the baseline output, and the output after RLIR post-training. No ground truth is required for visual quality cases. The key regions in the image are marked with red boxes. Post-training with RLIR mitigates action inconsistencies and image blurring.}
    \label{fig: maincase}
\end{figure*}
\subsection{Qualitative Analysis}

Figure~\ref{fig: maincase} presents qualitative comparisons of generations before and after RLIR post-training; additional examples are given in Appendix~\ref{appendix:more_case}. In the first case, the baseline model fails to accurately depict the hand’s digging action, yielding a mismatch between excavation progress. In the second case, the baseline shows limited fine-grained distance perception, causing a noticeable misalignment of the character’s position with the ground truth. In the final case, under rapid movement, the baseline produces localized pixel blur. By contrast, the RLIR-post-trained model effectively resolves these issues, in line with the improvement in quantitative results.

\begin{figure*}[!thbp]
    \centering
    \includegraphics[width=0.98\linewidth]{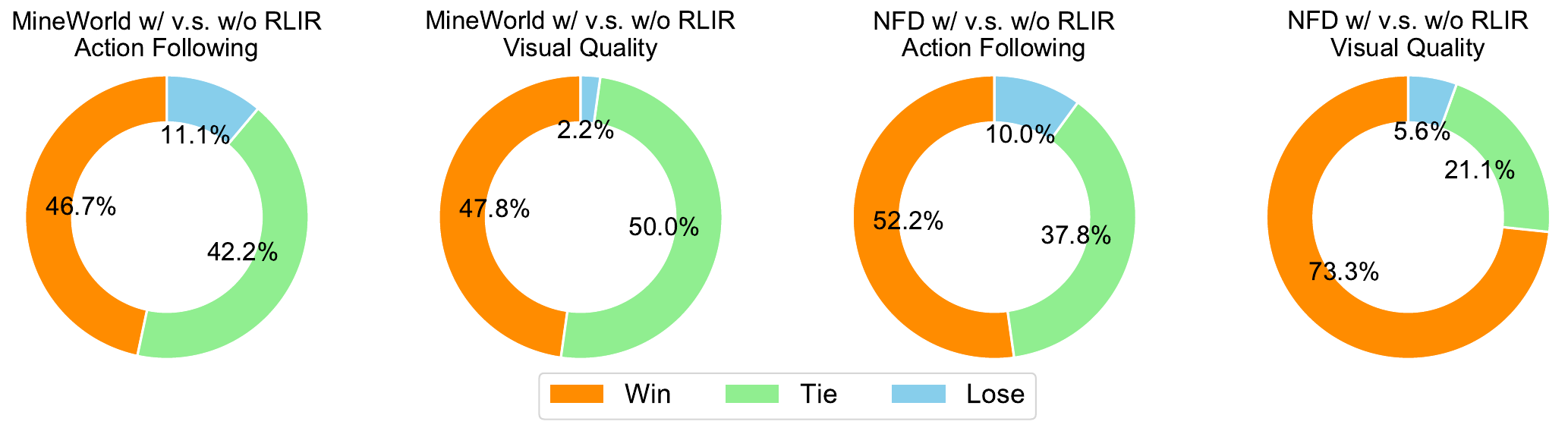}
    \caption{Human evaluation results for MineWorld and NFD with or without RLIR post-training. ``Win'' indicates the post-training results outperform the original one, while ``Lose'' represents the opposite. The results demonstrate that RLIR post-training yields higher human preference ratings for both visual quality and action-following ability.}
    \label{fig: human_evaluation}
\end{figure*}

We also conduct human evaluation as a complement to automatic metrics. For both the autoregressive world model and the diffusion world model, we randomly sample 10 videos from the evaluation set. Evaluators score each output along two dimensions: action-following ability and visual quality. The corresponding ground-truth videos are provided as references for judging action-following. As shown in Figure~\ref{fig: human_evaluation}, the models post-trained with RLIR exhibit a clear and consistent improvement over their counterparts on both criteria.

\section{Analysis}
\subsection{Different Reward Functions}
We evaluate the effectiveness of Reinforcement Learning with
Inverse Rewards (RLIR) by comparing it with human preference reward (e.g., VideoAlign~\citep{videoalign}) and pixel-level verifiable reward proposed in RLVR-World~\citep{rlvrworld}. VideoAlign uses 180k human-preference annotations to train a reward model that assigns separate scores to visual quality, motion dynamics, and text alignment. Since the world model is not text-conditioned, we use the mean of the first two dimensions as the reward. In contrast, RLVR-World treats the ground truth video directly as a verifiable reward signal. Concretely, its reward is the sum of the $L_1$ loss and the perceptual loss ($\mathrm{LPIPS}$) between the predictions and the ground-truth frames, $x_i$ means the i-th frame in the video:

\begin{equation}
\label{equ:rlvrworld}
R_{T_j} = - \sum_{i=1}^{n} \left[ L_{1}\big(\hat{x}_{i}, {x}_{i}\big) + \mathrm{LPIPS}\big(\hat{x}_{i}, {x}_{i}\big) \right]
\end{equation}

\begin{figure}[htbp]
  \centering
  \begin{minipage}[t]{0.38\textwidth}
   \vspace{0pt}
    \centering
    \includegraphics[width=\linewidth]{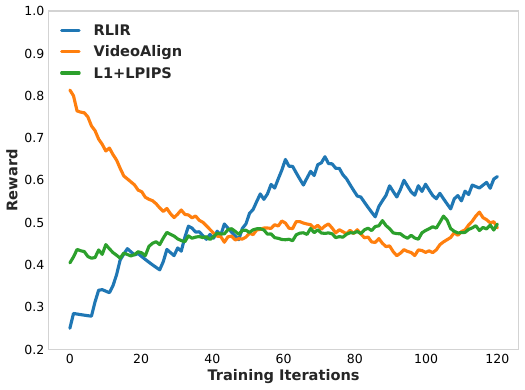}
    \captionof{figure}{Reward curves for VideoAlign, L1+LPIPS, and RLIR. We rescale the rewards to range (0, 1) for representation.}
    \label{fig:reward_curves}
  \end{minipage}\hfill
  \begin{minipage}[t]{0.58\textwidth}
    \vspace{2pt}
    \centering
    \small
    \resizebox{\textwidth}{!}{
        \begin{tabular}{cl|cccc}
            \toprule
            Model & Method  & F1$\uparrow$   & FVD$\downarrow$        & PSNR$\uparrow$      & IQ$\uparrow$ \\ 
            \midrule
            \multirow{4}{*}{\textbf{\makecell[c]{Mine-\\World}}}
            & Baseline     & 0.70        & 231       & 15.32     &  0.677             \\
            & \cellcolor{lightgreen} w/ $L_{1}+\mathrm{LPIPS}$      &\cellcolor{lightgreen} 0.71      & \cellcolor{lightgreen}228       & \cellcolor{lightgreen}15.47     &  \cellcolor{lightgreen}0.673               \\
            & \cellcolor{lightorange} w/ VideoAlign     & \cellcolor{lightorange}0.73    & \cellcolor{lightorange}219      &  \cellcolor{lightorange}15.50      & \cellcolor{lightorange}0.669                  \\
            & \cellcolor{lightblue} w/ RLIR   & \cellcolor{lightblue}\textbf{0.81}   & \cellcolor{lightblue}\textbf{207}     & \cellcolor{lightblue}\textbf{15.78}     &  \cellcolor{lightblue}\textbf{0.678}               \\
            \midrule
            \multirow{4}{*}{\textbf{NFD}} 
            & Baseline   & 0.77      & 184     & 16.95     &   0.692             \\
            & \cellcolor{lightgreen} w/ $L_{1}+\mathrm{LPIPS}$    &      \cellcolor{lightgreen}0.77  & \cellcolor{lightgreen}193  & \cellcolor{lightgreen}17.09   &   \cellcolor{lightgreen}0.645             \\
            & \cellcolor{lightorange} w/ VideoAlign     &  \cellcolor{lightorange}0.76      &  \cellcolor{lightorange}181       & \cellcolor{lightorange}17.45     &  \cellcolor{lightorange}\textbf{0.689 }          \\
            & \cellcolor{lightblue} w/ RLIR   & \cellcolor{lightblue}\textbf{0.83}      &  \cellcolor{lightblue}\textbf{180}   & \cellcolor{lightblue}\textbf{17.48}       &   \cellcolor{lightblue}0.688             \\
            \bottomrule
        \end{tabular}
    }
    \captionof{table}{Performance differences among three methods on 700M-parameter models, IQ is short for ``image quality''. Pixel-level verifiable reward yields no consistent improvements across models, and the human-preference reward likewise fails to improve performance significantly.}
    \label{tab:analysis}
  \end{minipage}
\end{figure}

We apply both methods to MineWorld and NFD during post-training. The results and reward curves are shown in Table~\ref{tab:analysis} and Figure~\ref{fig:reward_curves}. The ineffectiveness of VideoAlign is straightforward to explain: human evaluators introduce substantial bias and noise, and subtle motion differences in videos are difficult for them to discern. For RLVR-World, we attribute the lack of gains to four main factors:
\begin{itemize}[leftmargin=1.8em]
    \item \textbf{Correlation with pre-training objectives.}
The pixel-level supervision provided by $L_{1}+\mathrm{LPIPS}$ is highly correlated with the pre-training objectives (cross-entropy loss for MineWorld or flow matching loss for NFD), which have already been optimized. Consequently, the policy initialization starts near a local optimum, restricting the effective exploration of RL.
    \item \textbf{Uniform weighting of all pixels.}
Both $L_{1}$ and $\mathrm{LPIPS}$ aggregate errors over the entire frame, implicitly treating all regions as equally important. For action-following evaluation, however, regions associated with the agent’s motion are more critical. In contrast, an Inverse Dynamics Model naturally allocates greater attention to action-relevant regions.
    \item \textbf{Conflict with the generative objective.}
In many settings, a world model must synthesize genuinely novel content (e.g., regions uncovered during agent exploration). In such cases, pixel-level rewards are ill-suited, as the newly generated areas lack a deterministic ground truth. The reward should instead emphasize the fidelity of controllable factors (e.g., the magnitude and direction of motion), rather than penalize non-unique viusal outputs.
    \item \textbf{Susceptible to reward hacking.} 
In our experiments, we find that the videos produced by RLVR-World exhibit an overall dark appearance. This is because part of the post-training dataset is relatively dark. Therefore, the model can inflate the reward by uniformly darkening frames rather than improving semantic or dynamical fidelity. In contrast, RLIR depends on the predicted action alignment and is largely invariant to global brightness shifts, making it more robust.
\end{itemize}

In addition, regarding reward granularity~\citep{goodteacher}, the human-preference model provides a coarse, video-level reward, whereas applying RLVR directly yields an overly fine, pixel-level signal. Both extremes are suboptimal. In contrast, RLIR offers a precise, semantically aligned frame-level reward that better matches the training requirements of world models.

\subsection{Ablation on Hyperparameters}
We perform separate ablation studies on the principal hyperparameters of the autoregressive world model and the diffusion world model. 

For MineWorld, we evaluate whether adding a KL penalty improves performance. We test across different model sizes and find that the KL penalty yields measurable gains only for small models. 

For NFD, we perform ablations on the number of denoising steps and the SDE noise level $\epsilon_t$. When the number of denoising steps increases to 40, performance improves slowly and marginally; the best results occur with 10–20 steps. Setting the noise level too low diminishes gains, while performance remains similar for noise levels between 0.5 and 0.75. Ablation results appear in Appendix~\ref{appendix:ablation}.

\section{Conclusion}\label{sec: conclusion}
We introduce Reinforcement Learning with Inverse Reward (RLIR), a novel framework for world model post-training that transforms unverifiable videos into verifiable rewards. By leveraging the Inverse Dynamics Model (IDM) to convert videos into corresponding action sequences, we are able to measure the performance of the world model by the accuracy of predicted actions. This accuracy is then used as the verifiable reward in the reinforcement learning post-training process to optimize the world model. Experiments conducted on both autoregressive and diffusion world models demonstrate the effectiveness of the proposed method, achieving a 5\%–10\% improvement in action-following accuracy and enhancing visual quality as well.

\paragraph{Limitations} (1) As the IDM cannot achieve perfect accuracy, the attainable performance of our method is bounded by the IDM’s accuracy. (2) Constrained by computational and data resources, the largest model used in this work has only 1.2 billion parameters. As a result, the base model may fall short of the performance upper bound that RLIR could theoretically achieve.

\paragraph{Future Work} Future work will evaluate the scalability of RLIR on larger-scale world models and broaden its applications, including other world models and modalities beyond video.
\clearpage
\bibliography{ref}

\begin{thebibliography}{60}
\providecommand{\natexlab}[1]{#1}
\providecommand{\url}[1]{\texttt{#1}}
\expandafter\ifx\csname urlstyle\endcsname\relax
  \providecommand{\doi}[1]{doi: #1}\else
  \providecommand{\doi}{doi: \begingroup \urlstyle{rm}\Url}\fi

\bibitem[Agarwal et~al.(2025)Agarwal, Ali, Bala, Balaji, Barker, Cai, Chattopadhyay, Chen, Cui, Ding, et~al.]{cosmos}
Niket Agarwal, Arslan Ali, Maciej Bala, Yogesh Balaji, Erik Barker, Tiffany Cai, Prithvijit Chattopadhyay, Yongxin Chen, Yin Cui, Yifan Ding, et~al.
\newblock Cosmos world foundation model platform for physical ai.
\newblock \emph{arXiv preprint arXiv:2501.03575}, 2025.

\bibitem[Baker et~al.(2022)Baker, Akkaya, Zhokov, Huizinga, Tang, Ecoffet, Houghton, Sampedro, and Clune]{vpt}
Bowen Baker, Ilge Akkaya, Peter Zhokov, Joost Huizinga, Jie Tang, Adrien Ecoffet, Brandon Houghton, Raul Sampedro, and Jeff Clune.
\newblock Video pretraining (vpt): Learning to act by watching unlabeled online videos.
\newblock \emph{Advances in Neural Information Processing Systems}, 35:\penalty0 24639--24654, 2022.

\bibitem[Bar et~al.(2025)Bar, Zhou, Tran, Darrell, and LeCun]{bar2025navigation}
Amir Bar, Gaoyue Zhou, Danny Tran, Trevor Darrell, and Yann LeCun.
\newblock Navigation world models.
\newblock In \emph{Proceedings of the Computer Vision and Pattern Recognition Conference}, pp.\  15791--15801, 2025.

\bibitem[Black et~al.(2023)Black, Nakamoto, Atreya, Walke, Finn, Kumar, and Levine]{susie}
Kevin Black, Mitsuhiko Nakamoto, Pranav Atreya, Homer Walke, Chelsea Finn, Aviral Kumar, and Sergey Levine.
\newblock Zero-shot robotic manipulation with pretrained image-editing diffusion models.
\newblock \emph{arXiv preprint arXiv:2310.10639}, 2023.

\bibitem[Chen et~al.(2024{\natexlab{a}})Chen, Mart{\'\i}~Mons{\'o}, Du, Simchowitz, Tedrake, and Sitzmann]{diffusionforcing}
Boyuan Chen, Diego Mart{\'\i}~Mons{\'o}, Yilun Du, Max Simchowitz, Russ Tedrake, and Vincent Sitzmann.
\newblock Diffusion forcing: Next-token prediction meets full-sequence diffusion.
\newblock \emph{Advances in Neural Information Processing Systems}, 37:\penalty0 24081--24125, 2024{\natexlab{a}}.

\bibitem[Chen et~al.(2024{\natexlab{b}})Chen, Siarohin, Menapace, Deyneka, Chao, Jeon, Fang, Lee, Ren, Yang, et~al.]{panda70m}
Tsai-Shien Chen, Aliaksandr Siarohin, Willi Menapace, Ekaterina Deyneka, Hsiang-wei Chao, Byung~Eun Jeon, Yuwei Fang, Hsin-Ying Lee, Jian Ren, Ming-Hsuan Yang, et~al.
\newblock Panda-70m: Captioning 70m videos with multiple cross-modality teachers.
\newblock In \emph{Proceedings of the IEEE/CVF Conference on Computer Vision and Pattern Recognition}, pp.\  13320--13331, 2024{\natexlab{b}}.

\bibitem[Cheng et~al.(2025)Cheng, He, Xu, Guo, He, and Bian]{NFD}
Xinle Cheng, Tianyu He, Jiayi Xu, Junliang Guo, Di~He, and Jiang Bian.
\newblock Playing with transformer at 30+ fps via next-frame diffusion.
\newblock \emph{arXiv preprint arXiv:2506.01380}, 2025.

\bibitem[Du et~al.(2023)Du, Yang, Dai, Dai, Nachum, Tenenbaum, Schuurmans, and Abbeel]{unipi}
Yilun Du, Sherry Yang, Bo~Dai, Hanjun Dai, Ofir Nachum, Josh Tenenbaum, Dale Schuurmans, and Pieter Abbeel.
\newblock Learning universal policies via text-guided video generation.
\newblock \emph{Advances in neural information processing systems}, 36:\penalty0 9156--9172, 2023.

\bibitem[Google(2025)]{genie3}
Google.
\newblock Genie 3.
\newblock \url{https://deepmind.google/discover/blog/genie-3-a-new-frontier-for-world-models/}, 2025.

\bibitem[Grattafiori et~al.(2024)Grattafiori, Dubey, Jauhri, Pandey, Kadian, Al-Dahle, Letman, Mathur, Schelten, Vaughan, et~al.]{llama3}
Aaron Grattafiori, Abhimanyu Dubey, Abhinav Jauhri, Abhinav Pandey, Abhishek Kadian, Ahmad Al-Dahle, Aiesha Letman, Akhil Mathur, Alan Schelten, Alex Vaughan, et~al.
\newblock The llama 3 herd of models.
\newblock \emph{arXiv preprint arXiv:2407.21783}, 2024.

\bibitem[Gu et~al.(2025)Gu, Mao, and Shou]{gu2025long}
Yuchao Gu, Weijia Mao, and Mike~Zheng Shou.
\newblock Long-context autoregressive video modeling with next-frame prediction.
\newblock \emph{arXiv preprint arXiv:2503.19325}, 2025.

\bibitem[Guo et~al.(2025{\natexlab{a}})Guo, Yang, Zhang, Song, Zhang, Xu, Zhu, Ma, Wang, Bi, et~al.]{deepseekr1}
Daya Guo, Dejian Yang, Haowei Zhang, Junxiao Song, Ruoyu Zhang, Runxin Xu, Qihao Zhu, Shirong Ma, Peiyi Wang, Xiao Bi, et~al.
\newblock Deepseek-r1: Incentivizing reasoning capability in llms via reinforcement learning.
\newblock \emph{arXiv preprint arXiv:2501.12948}, 2025{\natexlab{a}}.

\bibitem[Guo et~al.(2025{\natexlab{b}})Guo, Ye, He, Wu, Jiang, Pearce, and Bian]{mineworld}
Junliang Guo, Yang Ye, Tianyu He, Haoyu Wu, Yushu Jiang, Tim Pearce, and Jiang Bian.
\newblock Mineworld: a real-time and open-source interactive world model on minecraft.
\newblock \emph{arXiv preprint arXiv:2504.08388}, 2025{\natexlab{b}}.

\bibitem[Ha \& Schmidhuber(2018)Ha and Schmidhuber]{worldmodel}
David Ha and J{\"u}rgen Schmidhuber.
\newblock World models.
\newblock \emph{arXiv preprint arXiv:1803.10122}, 2018.

\bibitem[Ho et~al.(2020)Ho, Jain, and Abbeel]{ho2020denoising}
Jonathan Ho, Ajay Jain, and Pieter Abbeel.
\newblock Denoising diffusion probabilistic models.
\newblock \emph{Advances in Neural Information Processing Systems (NeurIPS)}, 33:\penalty0 6840--6851, 2020.

\bibitem[Hore \& Ziou(2010)Hore and Ziou]{psnr}
Alain Hore and Djemel Ziou.
\newblock Image quality metrics: Psnr vs. ssim.
\newblock In \emph{2010 20th international conference on pattern recognition}, pp.\  2366--2369. IEEE, 2010.

\bibitem[Hu et~al.(2023)Hu, Russell, Yeo, Murez, Fedoseev, Kendall, Shotton, and Corrado]{hu2023gaia}
Anthony Hu, Lloyd Russell, Hudson Yeo, Zak Murez, George Fedoseev, Alex Kendall, Jamie Shotton, and Gianluca Corrado.
\newblock Gaia-1: A generative world model for autonomous driving.
\newblock \emph{arXiv preprint arXiv:2309.17080}, 2023.

\bibitem[Huang et~al.(2025)Huang, Zhou, Rabeti, Korovko, Ling, Ren, Shen, Gao, Slepichev, Lin, et~al.]{vipe}
Jiahui Huang, Qunjie Zhou, Hesam Rabeti, Aleksandr Korovko, Huan Ling, Xuanchi Ren, Tianchang Shen, Jun Gao, Dmitry Slepichev, Chen-Hsuan Lin, et~al.
\newblock Vipe: Video pose engine for 3d geometric perception.
\newblock \emph{arXiv preprint arXiv:2508.10934}, 2025.

\bibitem[Huang et~al.(2024)Huang, He, Yu, Zhang, Si, Jiang, Zhang, Wu, Jin, Chanpaisit, et~al.]{vbench}
Ziqi Huang, Yinan He, Jiashuo Yu, Fan Zhang, Chenyang Si, Yuming Jiang, Yuanhan Zhang, Tianxing Wu, Qingyang Jin, Nattapol Chanpaisit, et~al.
\newblock Vbench: Comprehensive benchmark suite for video generative models.
\newblock In \emph{Proceedings of the IEEE/CVF Conference on Computer Vision and Pattern Recognition}, pp.\  21807--21818, 2024.

\bibitem[Jaech et~al.(2024)Jaech, Kalai, Lerer, Richardson, El-Kishky, Low, Helyar, Madry, Beutel, Carney, et~al.]{openaio1}
Aaron Jaech, Adam Kalai, Adam Lerer, Adam Richardson, Ahmed El-Kishky, Aiden Low, Alec Helyar, Aleksander Madry, Alex Beutel, Alex Carney, et~al.
\newblock Openai o1 system card.
\newblock \emph{arXiv preprint arXiv:2412.16720}, 2024.

\bibitem[Ke et~al.(2021)Ke, Wang, Wang, Milanfar, and Yang]{musiq}
Junjie Ke, Qifei Wang, Yilin Wang, Peyman Milanfar, and Feng Yang.
\newblock Musiq: Multi-scale image quality transformer.
\newblock In \emph{Proceedings of the IEEE/CVF international conference on computer vision}, pp.\  5148--5157, 2021.

\bibitem[Lin et~al.(2023)Lin, Ye, Zhu, Cui, Ning, Jin, and Yuan]{videollava}
Bin Lin, Yang Ye, Bin Zhu, Jiaxi Cui, Munan Ning, Peng Jin, and Li~Yuan.
\newblock Video-llava: Learning united visual representation by alignment before projection.
\newblock \emph{arXiv preprint arXiv:2311.10122}, 2023.

\bibitem[Lin et~al.(2024{\natexlab{a}})Lin, Ge, Cheng, Li, Zhu, Wang, He, Ye, Yuan, Chen, et~al.]{opensoraplan}
Bin Lin, Yunyang Ge, Xinhua Cheng, Zongjian Li, Bin Zhu, Shaodong Wang, Xianyi He, Yang Ye, Shenghai Yuan, Liuhan Chen, et~al.
\newblock Open-sora plan: Open-source large video generation model.
\newblock \emph{arXiv preprint arXiv:2412.00131}, 2024{\natexlab{a}}.

\bibitem[Lin et~al.(2024{\natexlab{b}})Lin, Tang, Ye, Cui, Zhu, Jin, Huang, Zhang, Pang, Ning, et~al.]{moellava}
Bin Lin, Zhenyu Tang, Yang Ye, Jiaxi Cui, Bin Zhu, Peng Jin, Jinfa Huang, Junwu Zhang, Yatian Pang, Munan Ning, et~al.
\newblock Moe-llava: Mixture of experts for large vision-language models.
\newblock \emph{arXiv preprint arXiv:2401.15947}, 2024{\natexlab{b}}.

\bibitem[Liu et~al.(2024)Liu, Yan, Zaharia, and Abbeel]{liu2024world}
Hao Liu, Wilson Yan, Matei Zaharia, and Pieter Abbeel.
\newblock World model on million-length video and language with blockwise ringattention.
\newblock \emph{arXiv preprint arXiv:2402.08268}, 2024.

\bibitem[Liu et~al.(2025{\natexlab{a}})Liu, Liu, Liang, Li, Liu, Wang, Wan, Zhang, and Ouyang]{flowgrpo}
Jie Liu, Gongye Liu, Jiajun Liang, Yangguang Li, Jiaheng Liu, Xintao Wang, Pengfei Wan, Di~Zhang, and Wanli Ouyang.
\newblock Flow-grpo: Training flow matching models via online rl.
\newblock \emph{arXiv preprint arXiv:2505.05470}, 2025{\natexlab{a}}.

\bibitem[Liu et~al.(2025{\natexlab{b}})Liu, Liu, Liang, Yuan, Liu, Zheng, Wu, Wang, Qin, Xia, et~al.]{videoalign}
Jie Liu, Gongye Liu, Jiajun Liang, Ziyang Yuan, Xiaokun Liu, Mingwu Zheng, Xiele Wu, Qiulin Wang, Wenyu Qin, Menghan Xia, et~al.
\newblock Improving video generation with human feedback.
\newblock \emph{arXiv preprint arXiv:2501.13918}, 2025{\natexlab{b}}.

\bibitem[Lu et~al.(2025)Lu, Zhou, Bao, Chen, Li, and Zhu]{dpm}
Cheng Lu, Yuhao Zhou, Fan Bao, Jianfei Chen, Chongxuan Li, and Jun Zhu.
\newblock Dpm-solver++: Fast solver for guided sampling of diffusion probabilistic models.
\newblock \emph{Machine Intelligence Research}, pp.\  1--22, 2025.

\bibitem[Ouyang et~al.(2022)Ouyang, Wu, Jiang, Almeida, Wainwright, Mishkin, Zhang, Agarwal, Slama, Ray, et~al.]{rlhf}
Long Ouyang, Jeffrey Wu, Xu~Jiang, Diogo Almeida, Carroll Wainwright, Pamela Mishkin, Chong Zhang, Sandhini Agarwal, Katarina Slama, Alex Ray, et~al.
\newblock Training language models to follow instructions with human feedback.
\newblock \emph{Advances in neural information processing systems}, 35:\penalty0 27730--27744, 2022.

\bibitem[Peebles \& Xie(2023)Peebles and Xie]{peebles2023scalable}
William Peebles and Saining Xie.
\newblock Scalable diffusion models with transformers.
\newblock In \emph{Proceedings of the IEEE/CVF international conference on computer vision}, pp.\  4195--4205, 2023.

\bibitem[Radford et~al.(2021)Radford, Kim, Hallacy, Ramesh, Goh, Agarwal, Sastry, Askell, Mishkin, Clark, et~al.]{clip}
Alec Radford, Jong~Wook Kim, Chris Hallacy, Aditya Ramesh, Gabriel Goh, Sandhini Agarwal, Girish Sastry, Amanda Askell, Pamela Mishkin, Jack Clark, et~al.
\newblock Learning transferable visual models from natural language supervision.
\newblock In \emph{International conference on machine learning}, pp.\  8748--8763. PmLR, 2021.

\bibitem[Razin et~al.(2025)Razin, Wang, Strauss, Wei, Lee, and Arora]{goodteacher}
Noam Razin, Zixuan Wang, Hubert Strauss, Stanley Wei, Jason~D Lee, and Sanjeev Arora.
\newblock What makes a reward model a good teacher? an optimization perspective.
\newblock \emph{arXiv preprint arXiv:2503.15477}, 2025.

\bibitem[Rombach et~al.(2022)Rombach, Blattmann, Lorenz, Esser, and Ommer]{rombach2022high}
Robin Rombach, Andreas Blattmann, Dominik Lorenz, Patrick Esser, and Bj{\"o}rn Ommer.
\newblock High-resolution image synthesis with latent diffusion models.
\newblock In \emph{Proceedings of the IEEE/CVF Conference on Computer Vision and Pattern Recognition (CVPR)}, pp.\  10684--10695, 2022.

\bibitem[Schuhmann et~al.(2022)Schuhmann, Beaumont, Vencu, Gordon, Wightman, Cherti, Coombes, Katta, Mullis, Wortsman, et~al.]{aespredictor}
Christoph Schuhmann, Romain Beaumont, Richard Vencu, Cade Gordon, Ross Wightman, Mehdi Cherti, Theo Coombes, Aarush Katta, Clayton Mullis, Mitchell Wortsman, et~al.
\newblock Laion-5b: An open large-scale dataset for training next generation image-text models.
\newblock \emph{Advances in neural information processing systems}, 35:\penalty0 25278--25294, 2022.

\bibitem[Schulman et~al.(2017)Schulman, Wolski, Dhariwal, Radford, and Klimov]{ppo}
John Schulman, Filip Wolski, Prafulla Dhariwal, Alec Radford, and Oleg Klimov.
\newblock Proximal policy optimization algorithms.
\newblock \emph{arXiv preprint arXiv:1707.06347}, 2017.

\bibitem[Shao et~al.(2024)Shao, Wang, Zhu, Xu, Song, Bi, Zhang, Zhang, Li, Wu, et~al.]{deepseekmath}
Zhihong Shao, Peiyi Wang, Qihao Zhu, Runxin Xu, Junxiao Song, Xiao Bi, Haowei Zhang, Mingchuan Zhang, YK~Li, Y~Wu, et~al.
\newblock Deepseekmath: Pushing the limits of mathematical reasoning in open language models.
\newblock \emph{arXiv preprint arXiv:2402.03300}, 2024.

\bibitem[Shlens(2014)]{kl_div}
Jonathon Shlens.
\newblock Notes on kullback-leibler divergence and likelihood.
\newblock \emph{arXiv preprint arXiv:1404.2000}, 2014.

\bibitem[Shrivastava et~al.(2025)Shrivastava, Awadallah, Balachandran, Garg, Behl, and Papailiopoulos]{gfpo}
Vaishnavi Shrivastava, Ahmed Awadallah, Vidhisha Balachandran, Shivam Garg, Harkirat Behl, and Dimitris Papailiopoulos.
\newblock Sample more to think less: Group filtered policy optimization for concise reasoning.
\newblock \emph{arXiv preprint arXiv:2508.09726}, 2025.

\bibitem[Tan et~al.(2025)Tan, Feng, Mao, Huang, Liu, Hao, Su, and Zhu]{anypos}
Hengkai Tan, Yao Feng, Xinyi Mao, Shuhe Huang, Guodong Liu, Zhongkai Hao, Hang Su, and Jun Zhu.
\newblock Anypos: Automated task-agnostic actions for bimanual manipulation.
\newblock \emph{arXiv preprint arXiv:2507.12768}, 2025.

\bibitem[Team et~al.(2025)Team, Wang, Liu, Wu, Gu, Wang, Zuo, Huang, Li, Zhang, et~al.]{hunyuanworld}
HunyuanWorld Team, Zhenwei Wang, Yuhao Liu, Junta Wu, Zixiao Gu, Haoyuan Wang, Xuhui Zuo, Tianyu Huang, Wenhuan Li, Sheng Zhang, et~al.
\newblock Hunyuanworld 1.0: Generating immersive, explorable, and interactive 3d worlds from words or pixels.
\newblock \emph{arXiv preprint arXiv:2507.21809}, 2025.

\bibitem[Tot et~al.(2025)Tot, Ishida, Lemkhenter, Bignell, Choudhury, Lovett, Fran{\c{c}}a, de~Mendon{\c{c}}a, Gupta, Gehring, et~al.]{trajfollow}
Marko Tot, Shu Ishida, Abdelhak Lemkhenter, David Bignell, Pallavi Choudhury, Chris Lovett, Luis Fran{\c{c}}a, Matheus Ribeiro~Furtado de~Mendon{\c{c}}a, Tarun Gupta, Darren Gehring, et~al.
\newblock Adapting a world model for trajectory following in a 3d game.
\newblock \emph{arXiv preprint arXiv:2504.12299}, 2025.

\bibitem[Unterthiner et~al.(2018)Unterthiner, Van~Steenkiste, Kurach, Marinier, Michalski, and Gelly]{FVD}
Thomas Unterthiner, Sjoerd Van~Steenkiste, Karol Kurach, Raphael Marinier, Marcin Michalski, and Sylvain Gelly.
\newblock Towards accurate generative models of video: A new metric \& challenges.
\newblock \emph{arXiv preprint arXiv:1812.01717}, 2018.

\bibitem[Van Den~Oord et~al.(2017)Van Den~Oord, Vinyals, et~al.]{vqvae}
Aaron Van Den~Oord, Oriol Vinyals, et~al.
\newblock Neural discrete representation learning.
\newblock \emph{Advances in neural information processing systems}, 30, 2017.

\bibitem[Wen et~al.(2025)Wen, Liu, Zheng, Xu, Ye, Wu, Liang, Wang, Li, Miao, et~al.]{rlvr}
Xumeng Wen, Zihan Liu, Shun Zheng, Zhijian Xu, Shengyu Ye, Zhirong Wu, Xiao Liang, Yang Wang, Junjie Li, Ziming Miao, et~al.
\newblock Reinforcement learning with verifiable rewards implicitly incentivizes correct reasoning in base llms.
\newblock \emph{arXiv preprint arXiv:2506.14245}, 2025.

\bibitem[Wu et~al.(2025{\natexlab{a}})Wu, Wu, He, Guo, Ye, Duan, and Bian]{wu2025geometry}
Haoyu Wu, Diankun Wu, Tianyu He, Junliang Guo, Yang Ye, Yueqi Duan, and Jiang Bian.
\newblock Geometry forcing: Marrying video diffusion and 3d representation for consistent world modeling.
\newblock \emph{arXiv preprint arXiv:2507.07982}, 2025{\natexlab{a}}.

\bibitem[Wu et~al.(2025{\natexlab{b}})Wu, Yin, Feng, and Long]{rlvrworld}
Jialong Wu, Shaofeng Yin, Ningya Feng, and Mingsheng Long.
\newblock Rlvr-world: Training world models with reinforcement learning.
\newblock \emph{arXiv preprint arXiv:2505.13934}, 2025{\natexlab{b}}.

\bibitem[Wu et~al.(2025{\natexlab{c}})Wu, Yang, Po, Xu, Liu, Lin, and Wetzstein]{wu2025video}
Tong Wu, Shuai Yang, Ryan Po, Yinghao Xu, Ziwei Liu, Dahua Lin, and Gordon Wetzstein.
\newblock Video world models with long-term spatial memory.
\newblock \emph{arXiv preprint arXiv:2506.05284}, 2025{\natexlab{c}}.

\bibitem[Wu et~al.(2023)Wu, Hao, Sun, Chen, Zhu, Zhao, and Li]{hps}
Xiaoshi Wu, Yiming Hao, Keqiang Sun, Yixiong Chen, Feng Zhu, Rui Zhao, and Hongsheng Li.
\newblock Human preference score v2: A solid benchmark for evaluating human preferences of text-to-image synthesis.
\newblock \emph{arXiv preprint arXiv:2306.09341}, 2023.

\bibitem[Xiao et~al.(2025)Xiao, Lan, Zhou, Ouyang, Yang, Zeng, and Pan]{xiao2025worldmem}
Zeqi Xiao, Yushi Lan, Yifan Zhou, Wenqi Ouyang, Shuai Yang, Yanhong Zeng, and Xingang Pan.
\newblock Worldmem: Long-term consistent world simulation with memory.
\newblock \emph{arXiv preprint arXiv:2504.12369}, 2025.

\bibitem[Xue et~al.(2025)Xue, Wu, Gao, Kong, Zhu, Chen, Liu, Liu, Guo, Huang, et~al.]{dancegrpo}
Zeyue Xue, Jie Wu, Yu~Gao, Fangyuan Kong, Lingting Zhu, Mengzhao Chen, Zhiheng Liu, Wei Liu, Qiushan Guo, Weilin Huang, et~al.
\newblock Dancegrpo: Unleashing grpo on visual generation.
\newblock \emph{arXiv preprint arXiv:2505.07818}, 2025.

\bibitem[Yang et~al.(2025)Yang, Niu, Liu, Ye, Lin, and Yuan]{lookback}
Shuo Yang, Yuwei Niu, Yuyang Liu, Yang Ye, Bin Lin, and Li~Yuan.
\newblock Look-back: Implicit visual re-focusing in mllm reasoning.
\newblock \emph{arXiv preprint arXiv:2507.03019}, 2025.

\bibitem[Ye et~al.(2025{\natexlab{a}})Ye, Guo, Wu, He, Pearce, Rashid, Hofmann, and Bian]{diagd}
Yang Ye, Junliang Guo, Haoyu Wu, Tianyu He, Tim Pearce, Tabish Rashid, Katja Hofmann, and Jiang Bian.
\newblock Fast autoregressive video generation with diagonal decoding.
\newblock \emph{arXiv preprint arXiv:2503.14070}, 2025{\natexlab{a}}.

\bibitem[Ye et~al.(2025{\natexlab{b}})Ye, He, Li, Lin, Yuan, Yan, Hou, and Yuan]{imgedit}
Yang Ye, Xianyi He, Zongjian Li, Bin Lin, Shenghai Yuan, Zhiyuan Yan, Bohan Hou, and Li~Yuan.
\newblock Imgedit: A unified image editing dataset and benchmark.
\newblock \emph{arXiv preprint arXiv:2505.20275}, 2025{\natexlab{b}}.

\bibitem[Yu et~al.(2025{\natexlab{a}})Yu, Qin, Wang, Wan, Zhang, and Liu]{gamefactory}
Jiwen Yu, Yiran Qin, Xintao Wang, Pengfei Wan, Di~Zhang, and Xihui Liu.
\newblock Gamefactory: Creating new games with generative interactive videos.
\newblock \emph{arXiv preprint arXiv:2501.08325}, 2025{\natexlab{a}}.

\bibitem[Yu et~al.(2025{\natexlab{b}})Yu, Zhang, Zhu, Yuan, Zuo, Yue, Dai, Fan, Liu, Liu, et~al.]{dapo}
Qiying Yu, Zheng Zhang, Ruofei Zhu, Yufeng Yuan, Xiaochen Zuo, Yu~Yue, Weinan Dai, Tiantian Fan, Gaohong Liu, Lingjun Liu, et~al.
\newblock Dapo: An open-source llm reinforcement learning system at scale.
\newblock \emph{arXiv preprint arXiv:2503.14476}, 2025{\natexlab{b}}.

\bibitem[Yuan et~al.(2025{\natexlab{a}})Yuan, He, Deng, Ye, Huang, Lin, Luo, and Yuan]{opens2v}
Shenghai Yuan, Xianyi He, Yufan Deng, Yang Ye, Jinfa Huang, Bin Lin, Jiebo Luo, and Li~Yuan.
\newblock Opens2v-nexus: A detailed benchmark and million-scale dataset for subject-to-video generation.
\newblock \emph{arXiv preprint arXiv:2505.20292}, 2025{\natexlab{a}}.

\bibitem[Yuan et~al.(2025{\natexlab{b}})Yuan, Huang, He, Ge, Shi, Chen, Luo, and Yuan]{consisid}
Shenghai Yuan, Jinfa Huang, Xianyi He, Yunyang Ge, Yujun Shi, Liuhan Chen, Jiebo Luo, and Li~Yuan.
\newblock Identity-preserving text-to-video generation by frequency decomposition.
\newblock In \emph{Proceedings of the Computer Vision and Pattern Recognition Conference}, pp.\  12978--12988, 2025{\natexlab{b}}.

\bibitem[Zhang \& Agrawala(2025)Zhang and Agrawala]{zhang2025packing}
Lvmin Zhang and Maneesh Agrawala.
\newblock Packing input frame context in next-frame prediction models for video generation.
\newblock \emph{arXiv preprint arXiv:2504.12626}, 2025.

\bibitem[Zheng et~al.(2025)Zheng, Liu, Li, Chen, Yu, Gao, Dang, Liu, Men, Yang, et~al.]{gspo}
Chujie Zheng, Shixuan Liu, Mingze Li, Xiong-Hui Chen, Bowen Yu, Chang Gao, Kai Dang, Yuqiong Liu, Rui Men, An~Yang, et~al.
\newblock Group sequence policy optimization.
\newblock \emph{arXiv preprint arXiv:2507.18071}, 2025.

\bibitem[Zhou et~al.(2024)Zhou, Pan, LeCun, and Pinto]{dinowm}
Gaoyue Zhou, Hengkai Pan, Yann LeCun, and Lerrel Pinto.
\newblock Dino-wm: World models on pre-trained visual features enable zero-shot planning.
\newblock \emph{arXiv preprint arXiv:2411.04983}, 2024.

\end{thebibliography}
\bibliographystyle{iclr2026_conference}
\newpage
\appendix
\setcounter{page}{1}

\startcontents[chapters]
\section*{\textbf{{Appendix}} }

\printcontents[chapters]{}{1}{}

\section{More Implementation Details}~\label{appendix:impl_detail}

\subsection{Details of Evaluation Metrics}
The Imaging Quality metric in VBench primarily evaluates low-level distortions in generated video frames (e.g., overexposure, noise, blur). VBench uses the MUSIQ~\cite{musiq} image-quality predictor, which can accommodate variable aspect ratios and resolutions. Each per-frame score (originally in the range 0–100) is divided by 100 to map it to [0,1], and the final metric is the arithmetic mean of the normalized scores across all frames in the video.

For action following metrics, actions in Minecraft can be grouped into $9$ classes, where $7$ of them represent discrete action classes and the other $2$ represent camera movement angles. For discrete classes, each one of them contains two or three exclusive actions such as $(\texttt{forward}, \texttt{backward})$ and $(\texttt{left}, \texttt{right})$. We provide the full grouping results in Table~\ref{tab:subtask}. Then, by taking the provided action as the ground truth and the predicted action from IDM as the prediction, we can utilize commonly utilized classification metrics including precision, recall and F1 score to evaluate the classification accuracy. We report both the macro scores to reduce the effect of imbalanced labels.

\begin{table}[htbp]
\centering
\caption{Classification Tasks and Their Labels}
\vspace{2mm}
\begin{tabular}{ l|c|c }
\toprule[1pt]
\textbf{Task Type} & \textbf{Actions} & \textbf{Labels} \\
\midrule
\multirow{3}{*}{Triple Classification} & \texttt{forward}, \texttt{backward} & \texttt{forward}, \texttt{backward}, \texttt{null} \\
& \texttt{left}, \texttt{right} & \texttt{left}, \texttt{right}, \texttt{null} \\
& \texttt{sprint}, \texttt{sneak} & \texttt{sprint}, \texttt{sneak}, \texttt{null} \\
\midrule
\multirow{4}{*}{Binary Classification} & \texttt{use} & \texttt{use}, \texttt{null} \\
& \texttt{attack} & \texttt{attack}, \texttt{null} \\
& \texttt{jump} & \texttt{jump}, \texttt{null} \\
& \texttt{drop} & \texttt{drop}, \texttt{null} \\
\bottomrule
\end{tabular}
\label{tab:subtask}
\end{table}

\subsection{Model Configuratons}
\paragraph{MineWorld}
We apply the proposed algorithm and post-train three MineWorld models of different sizes—300M, 700M, and 1.2B parameters—based on the LLaMA architecture. The base model configurations are summarized in Table~\ref{tab:model_arch_ar}.

\begin{table}[htbp]
    \centering
    \caption{The configuration of different size of MineWorld models.}
    \begin{tabular}{l|c|c|c|c}
        \toprule[1pt]
            & \textbf{Hidden Dim.} & \textbf{MLP Dim.} & \textbf{Num. Heads} & \textbf{Num. Layers} \\
        \midrule
        300M &  1024 & 4096 & 16 & 20  \\
        700M &  2048 & 4096 & 32 & 20 \\
        1.2B &  2048 & 8192 & 32 & 20 \\
        \bottomrule[1pt]
    \end{tabular}
    \label{tab:model_arch_ar}
\end{table}

\paragraph{NFD}
We post train on 300M and 770M parameter NFD models. Their base configurations are summarized in Table~\ref{tab:model_arch_df}. The NFD architecture comprises Diffusion Transformer Blocks. 

NFD employs an image-level tokenizer to transform each frame into a sequence of latents to enable the frame-level interaction with the model. For actions, NFD quantizes camera angles into discrete bins, and categorize other actions into 7 exclusive classes, each represented by a unique token.

NFD leverages a Block-wise Causal Attention mechanism that combines bidirectional attention within each frame and causal dependencies across frames to model spatio-temporal dependencies efficiently. For each token in a frame, it will attend to all tokens within the same frame (i.e., intra-frame attention), as well as to all tokens in preceding frames (i.e., causal inter-frame attention). 

NFD utilizes a linear layer to map the actions into action vectors and adopt adaLN-zero conditioning.

\begin{table}[htbp]
    \centering
    \caption{The configuration of different size of NFD models.}
    \begin{tabular}{l|c|c|c|c}
        \toprule[1pt]
            & \textbf{Hidden Dim.} & \textbf{MLP Dim.} & \textbf{Num. Heads} & \textbf{Num. Layers} \\
        \midrule
        310M &  1024 & 2730 & 16 & 16  \\
        774M &  1536 & 4096 & 24 & 18 \\
        \bottomrule[1pt]
    \end{tabular}
    \label{tab:model_arch_df}
\end{table}

\subsection{Experimental Settings}
\paragraph{MineWorld}
Table~\ref{tab:hyperparams_ar} lists the hyperparameters used for MineWorld post-training. 

\begin{table}[htbp]
    \centering
    \caption{Hyper-parameters for MineWorld.}
    \begin{tabular}{c|c}
        \toprule[1pt]
       \textbf{Hyperparameter}  &  \textbf{Value} \\
       \midrule
       Learning rate scheduler   &  cosine \\
       Learning rate & $3e^{-5}$ \\
       Optimizer & AdamW \\
       Rollout &  24   \\
       Clip Ratio  & 0.2 \\
       Samples per iteration & 32 \\
       \bottomrule[1pt]
    \end{tabular}
    \label{tab:hyperparams_ar}
\end{table}

\paragraph{NFD}
Table~\ref{tab:hyperparams_df} summarizes the hyperparameters used for NFD post-training.

\begin{table}[htbp]
    \centering
    \caption{Hyper-parameters for NFD.}
    \begin{tabular}{c|c}
        \toprule[1pt]
       \textbf{Hyperparameter}  &  \textbf{Value} \\
       \midrule
       Learning rate scheduler   &  cosine \\
       Learning rate & $1e^{-5}$ \\
       Optimizer & AdamW \\
       Rollout &  24   \\
       Clip Ratio  & 0.2 \\
       Samples per iteration & 16  \\
       Sampling steps & 10   \\
       Noise level $\epsilon_t$ & 0.75 \\
       Timestep Selection $\tau$  & 0.6 \\
       \bottomrule[1pt]
    \end{tabular}
    \label{tab:hyperparams_df}
\end{table}

\section{Ablation Studies}~\label{appendix:ablation}
\subsection{MineWorld}
For the autoregressive world models, we investigate the effect of introducing a KL-divergence constraint. The KL penalty term is used to regulate the divergence between the online policy and the frozen reference policy, the goal of KL-divergence is to align the model behavior without diverging too far from the initial model. With a fixed KL penalty of 1e-4 across model sizes, smaller models benefit (better action following and visual quality), whereas larger models suffer performance drops. We show the effectiveness of the KL penalty across all models in Table~\ref{tab: ablation_ar}.

\begin{table}[htbp]
    \centering
    \caption{Ablation study on kl penalty. the kl penalty coefficient is set to $\beta=1e-4$.}
    \vspace{1mm}
    \label{tab: ablation_ar}
    \resizebox{\textwidth}{!}{
        \begin{tabular}{l|ccccccc}
            \toprule
             Model  & F1$\uparrow$       & Recall$\uparrow$  & Precision$\uparrow$  & FVD$\downarrow$        & PSNR$\uparrow$      & Img. Qual.$\uparrow$  & Dynamic\\ 
            \midrule
             \rowcolor{lightblue} 300M w/ $kl$   & 0.77     & 0.76      & 0.79     & 231    & 15.58          & 0.672 &   0.97    \\
            300M w/o $kl$    & 0.69       & 0.69      & 0.73      & 231       & 15.65     & 0.672  & 0.98     \\
            
            \midrule
            
             700M w/ $kl$    & 0.73       & 0.73     & 0.75      & 210       & 15.91  & 0.678     & 0.98              \\
             \rowcolor{lightblue} 700M w/o $kl$   & 0.81     & 0.80       & 0.84     & 207     & 15.78       & 0.678 & 0.97        \\
            \midrule
             1200M w/ $kl$    & 0.80      & 0.79      & 0.82      & 219       & 15.80      &  0.683 &  0.97          \\
             \rowcolor{lightblue} 1200M w/o $kl$  & 0.81      & 0.81     & 0.83   & 205     & 15.99       &  0.684  & 0.96          \\ 
            \bottomrule
        \end{tabular}
    }
\end{table}

\subsection{NFD}
To investigate the impact of different timesteps on optimization, we keep other hyperparameters constant and test 10, 20, and 40 steps. When the number of denoising steps is increased to 40, performance improved only slowly and marginally; the best results were observed with 10–20 steps.
Regarding noise level, we test 0.25, 0.5 and 0.75. Setting a too low noise level diminishes performance gains, while results remain similar for noise levels between 0.5 and 0.75.
Table~\ref{tab: ablation_df_step} and~\ref{tab: ablation_df_noise} shows the effect of denoising steps and noise level.

\begin{table}[htbp]
    \centering
    \caption{Ablation study of NFD on denoising steps.}
    \vspace{1mm}
    \label{tab: ablation_df_step}
    \resizebox{\textwidth}{!}{
        \begin{tabular}{lc|ccccccc}
            \toprule
             Model & Step  & F1$\uparrow$       & Recall$\uparrow$  & Precision$\uparrow$  & FVD$\downarrow$        & PSNR$\uparrow$      & Img. Qual.$\uparrow$  & Dynamic\\ 
            \midrule
             \multirow{3}{*}{\textbf{310M}} & \cellcolor{lightblue} 10   & \cellcolor{lightblue}0.76     &\cellcolor{lightblue} 0.76      & \cellcolor{lightblue}0.77     & \cellcolor{lightblue}195    & \cellcolor{lightblue}17.38          & \cellcolor{lightblue}0.687 &  \cellcolor{lightblue} 0.99    \\
            & 20    & 0.74       & 0.74      & 0.76      & 186       & 17.23     & 0.687 &  1.00       \\
            & 40    & 0.70      & 0.70      & 0.74      & 221       & 16.72     & 0.667   & 1.00     \\
            
            \midrule
            
             \multirow{3}{*}{\textbf{774M}} & \cellcolor{lightblue} 10   & \cellcolor{lightblue}0.83     & \cellcolor{lightblue}0.83      & \cellcolor{lightblue}0.85     & \cellcolor{lightblue}180    & \cellcolor{lightblue}17.48          & \cellcolor{lightblue}0.688 &   \cellcolor{lightblue}1.00    \\
            & 20    & 0.81       & 0.84      & 0.80      & 185       & 17.35     & 0.683    &   1.00     \\
            & 40    & 0.77      & 0.78      & 0.78      & 180       & 17.47     & 0.686      &   1.00     \\
            \bottomrule
        \end{tabular}
    }
\end{table}

\begin{table}[htbp]
    \centering
    \caption{Ablation study of NFD on noise level.}
    \vspace{1mm}
    \label{tab: ablation_df_noise}
    \resizebox{\textwidth}{!}{
        \begin{tabular}{lc|ccccccc}
            \toprule
             Model & \makecell[c]{Noise\\Level ($\epsilon_{t}$)}  & F1$\uparrow$       & Recall$\uparrow$  & Precision$\uparrow$  & FVD$\downarrow$        & PSNR$\uparrow$      & Img. Qual.$\uparrow$  & Dynamic\\ 
            \midrule
             \multirow{3}{*}{\textbf{310M}} & \cellcolor{lightblue} 0.75   & \cellcolor{lightblue}0.76     &\cellcolor{lightblue} 0.76      & \cellcolor{lightblue}0.77     & \cellcolor{lightblue}195    & \cellcolor{lightblue}17.38          & \cellcolor{lightblue}0.687 &  \cellcolor{lightblue} 0.99    \\
            & 0.50    & 0.75       & 0.75      & 0.76      & 196       & 17.39     & 0.684     & 0.99     \\
            & 0.25    & 0.76      & 0.76      & 0.77      & 199       & 17.35     & 0.687  &  0.99     \\
            
            \midrule
            
             \multirow{3}{*}{\textbf{774M}} & \cellcolor{lightblue} 0.75   & \cellcolor{lightblue}0.83     & \cellcolor{lightblue}0.83      & \cellcolor{lightblue}0.85     & \cellcolor{lightblue}180    & \cellcolor{lightblue}17.48          & \cellcolor{lightblue}0.688 &   \cellcolor{lightblue}1.00    \\
            & 0.50    & 0.83       & 0.83      & 0.84      & 187       & 17.40     & 0.683     & 0.99        \\
            & 0.25   & 0.83       & 0.83      & 0.84      & 183      &    17.43  & 0.684      & 1.00       \\
            \bottomrule
        \end{tabular}
    }
\end{table}

\section{More Visualization Results}\label{appendix:more_case}
\begin{figure*}[!htbp]
    \centering
    \includegraphics[width=1\linewidth]{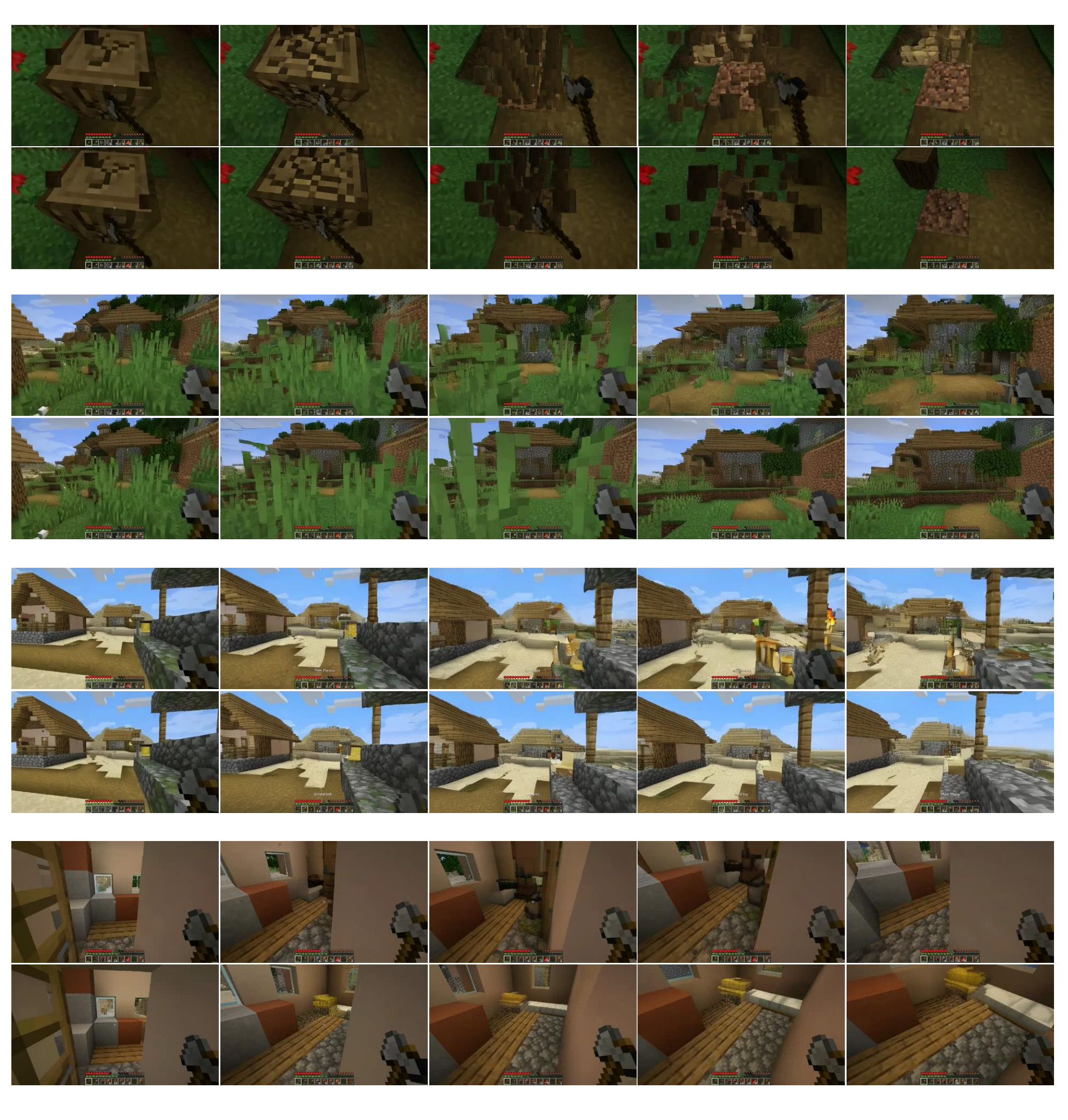}
    \caption{More Qualitative results. The top row displays the baseline output, and the bottom row is the output after post-training.}
    \label{fig: morecase}
\end{figure*}

We present additional visualization cases in Figure~\ref{fig: morecase}: the first three cases are from NFD, and the last case is from MineWorld. More videos can be found in the supplementary material.

\end{document}